# Temporal Contrastive Transformer for Financial Crime Detection: Self-Supervised Sequence Embeddings via Predictive Contrastive Coding

Danny Butvinik, *Chief Data Scientist, NICE Actimize*
Yonit Marcus, *Lead Data Scientist, NICE Actimize*
Nitzan Tal, *Senior Data Scientist, NICE Actimize*
Gabrielle Azoulay, *Senior Data Scientist, NICE Actimize*

## ABSTRACT

We introduce the Temporal Contrastive Transformer (TCT), a representation learning framework designed to capture contextual temporal dynamics in sequences of financial transactions. The model is trained using a self-supervised contrastive objective to produce embeddings that encode behavioral patterns over time, with the goal of supporting downstream fraud detection tasks. We evaluate TCT in a realistic setting by using the learned embeddings as input features to a gradient boosting classifier. Experimental results show that embeddings alone achieve meaningful predictive performance (AUC 0.8644), indicating that the model captures non-trivial temporal structure. However, when combined with domain-engineered features, no measurable improvement is observed over the baseline (AUC 0.9205 vs. 0.9245), suggesting that the learned representations largely overlap with existing feature abstractions. These findings position TCT as a promising representation learning approach that captures relevant behavioral signal, while highlighting the challenges of achieving additive value over strong domain features. The results reflect an intermediate stage in the development of temporal representation learning for financial crime detection and motivate further research on model architecture, training objectives, and integration strategies. At this early stage, achieving performance comparable to a strong feature-engineered baseline is itself a meaningful outcome, indicating that learned representations approximate domain-specific features without manual engineering. While not yet production-ready, these results point to a promising direction for reducing reliance on feature engineering in financial crime detection.

## 1. INTRODUCTION

Financial institutions operate under a dual mandate in transaction monitoring: detecting payment fraud, where adversaries exploit real-time payment systems to misappropriate funds and identify patterns indicative of money laundering, where seemingly legitimate transaction flows conceal illicit activity. Although both problems share a common structural nature, anomalous behavior expressed across sequences of financial events; they are typically addressed by separate rule-based or supervised machine learning systems that evaluate transactions largely in isolation.

Classical approaches rely on hand-crafted features and threshold-based rules derived from domain expertise. While effective, these methods are inherently limited in their ability to generalize evolving adversarial patterns and require continuous manual adaptation. Recent advances in deep learning suggest that temporal context is essential: both fraud and laundering behaviors manifest as structured, sequential patterns unfolding over time rather than as isolated anomalies [1, 2].

Despite progress in sequence modeling, two fundamental challenges remain. First, labeled data in both fraud and AML settings are scarce and highly imbalanced, constraining the effectiveness of fully supervised approaches and increasing the risk of overfitting. Second, transaction data are intrinsically heterogeneous, combining categorical identifiers, continuous monetary values, temporal intervals, and derived behavioral aggregates, which complicates the construction of unified representations.

To address these challenges, we propose the Temporal Contrastive Transformer (TCT), a pre-training framework that learns mappings from transaction sequences to compact embedding vectors using unlabeled data. TCT integrates two complementary inductive biases. The Temporal Fusion Transformer (TFT) component [3] provides a structured architecture for multi-horizon temporal modeling, enabling separation of static and dynamic inputs, adaptive variable selection, and hierarchical aggregation of temporal dependencies. The Contrastive Predictive Coding (CPC) objective [4] trains the encoder to capture predictive temporal structure by contrasting future latent representations against negative samples, thereby encoding behavioral dynamics without reliance on labels.

The learned embeddings are intended to capture latent temporal patterns that are not explicitly encoded in raw features. In this study, we evaluate their utility in a downstream fraud detection task by combining them with domain-engineered behavioral features. Empirical results show that while the embeddings encode meaningful signal, they do not yield measurable improvement over strong feature-based baselines, indicating substantial overlap between learned and engineered representations. The central hypothesis is that self-supervised temporal representations are hypothesized to provide incremental predictive value beyond domain-engineered features. The objective of this work is not to demonstrate immediate performance gains, but to evaluate whether self-supervised temporal representations can approximate the predictive power of domain-engineered features. Establishing such parity at an early stage would indicate a viable direction for reducing reliance on manual feature engineering.

The remainder of the paper is organized as follows. Section 2 reviews related work. Section 3 presents the TCT framework. Section 4 describes the experimental setup. Section 5 reports results. Section 6 discusses implications and limitations. Section 7 concludes.

## 2. RELATED WORK

### *2.1 Temporal Sequence Models for Fraud and AML Detection*

Recurrent neural networks and their gated variants LSTM [1] and GRU [2] have been applied to fraud detection by modeling transaction sequences as variable-length time series [24]. These approaches capture temporal dependencies and often improve recall of rare fraud events compared to instance-level classifiers. However, they rely on labeled data for supervision and typically couple representation learning with the classification objective, limiting transferability. Bidirectional extensions [21] enhance contextual encoding by incorporating information from both past and future time steps, while Transformer-based architectures [8] address the limitations of recurrent models in capturing long-range dependencies. Nevertheless, these models remain predominantly supervised and task-specific.

In the AML domain, sequence-based models have been explored for identifying structuring and layering behaviors [12], where suspicious activity unfolds across multiple temporally linked transactions. Despite these advances, most existing approaches are tailored to specific tasks and rely on labeled supervision. In contrast, TCT adopts a self-supervised formulation, aiming to learn a general-purpose temporal representation from unlabeled transaction sequences that can be reused across downstream tasks such as fraud detection and AML analytics.

### *2.2 Self-Supervised and Contrastive Representation Learning*

Self-supervised learning has become a central paradigm for representation learning across domains. In natural language processing, BERT [22] demonstrated that masked-token prediction on large unlabeled corpora yields transferable contextual embeddings. In temporal domains, Contrastive Predictive Coding (CPC) [4] introduced a predictive objective that learns representations by distinguishing true future latent states from negative samples. More broadly, contrastive learning frameworks such as MoCo [5] and SimCLR [6] have shown that representations can be learned by maximizing agreement between related views of the data without requiring labels.

The InfoNCE objective [7] formalizes this approach by providing a tractable lower bound on mutual information between paired representations, with connections to noise-contrastive estimation. TCT builds on these principles by defining contrastive tasks over temporal sub-sequences of financial transactions, encouraging the model to encode predictive behavioral structure inherent in sequential activity.

### *2.3 Temporal Fusion Transformer*

The Temporal Fusion Transformer (TFT) [3] was introduced for multi-horizon time-series forecasting and provides architectural components well-suited to heterogeneous sequential data. It incorporates a variable selection mechanism for adaptive feature gating, gated residual networks [14] with gated linear units and layer normalization [14, 15], and a multi-head self-attention mechanism [8] operating over temporally encoded representations. These elements enable the model to integrate static and dynamic inputs while capturing both short-term and long-term dependencies.

In TCT, the TFT architecture is repurposed from forecasting to representation learning. Specifically, the predictive regression head is replaced with a contrastive objective, allowing the model to learn latent temporal embeddings without requiring labeled targets.

### *2.4 Embedding-Based Financial Crime Detection*

Embedding-based approaches have been explored in financial crime detection through both graph and sequence paradigms. Graph neural networks (GNNs) [9, 13] generate entity-level embeddings by leveraging relational structures in transaction networks, while autoencoder-based methods [10] learn compressed representations of transactional behavior. The Elliptic dataset [23] has served as a benchmark for AML detection on blockchain graphs, demonstrating the value of incorporating multi-hop relational features for identifying laundering patterns.

These approaches highlight the importance of representation learning but often focus on either structural (graph-based) or transactional (sequence-based) aspects in isolation. TCT complements this landscape by learning temporally structured embeddings from transaction sequences, which can be used directly in downstream models or integrated as features within graph-based frameworks.

## 3. THE TEMPORAL CONTRASTIVE TRANSFORMER

Let $P$ denote a population of financial parties. For each party $p \in P$, define a time-ordered sequence of events $S_p = (e_{p,1}, e_{p,2}, \dots, e_{p,T})$, where each event $e_{p,t} \in \mathcal{X}$ is represented by a feature vector comprising numerical, categorical, temporal, and derived attributes. We partition the feature space into static features $\mathcal{X}_s \subset \mathcal{X}$ and dynamic features $\mathcal{X}_d \subset \mathcal{X}$. The objective of TCT is to learn an encoder $f_\theta: \mathcal{X}^L \to \mathbb{R}^d$, which maps a sequence of $L$ events to a compact embedding vector $e_p = f_\theta(S_p) \in \mathbb{R}^d$.

### *3.1 Feature Encoding and Variable Selection*

Numerical features are standardized, and categorical features with vocabulary size $V_j$ are embedded via matrices $E_j \in \mathbb{R}^{V_j \times d_e}$. Static and dynamic inputs are encoded through $\psi_s$ and $\psi_d$, yielding

$$\tilde{x}_t = [\psi_s(x_{s,t}); \psi_d(x_{d,t})].$$

A Variable Selection Network (VSN) [3] assigns soft importance weights. Let $\phi_t \in \mathbb{R}^{M \cdot d_e}$ be concatenated embeddings. Then:

$$v_t = \text{Softmax}(\text{GRN}_v(\phi_t, c_s)) \in \mathbb{R}^M$$

where $c_s$ is a static context vector. The GRN is defined as:

$$\eta_2 = \text{ELU}(W_2 a + W_3 c + b_2), \eta_1 = W_1 \eta_2 + b_1$$

$$\text{GRN}(a, c) = \text{LayerNorm}(a + \text{GLU}(\eta_1))$$

The selected representation is:

$$\xi_t = \sum_{j=1}^{M} v_{t,j} \; \Xi_{t,j}$$

### *3.2 Hierarchical Temporal Encoding*

Temporal dynamics are modeled using a hierarchical encoder composed of two stacked recurrent modules operating at different temporal resolutions.

The short-term encoder (Enc1), implemented as an LSTM [1] or GRU [2], processes local sub-sequences of length $L$. At each time step $t$, the hidden state is updated as:

$$h_t^{(1)} = \text{LSTM}(\xi_t, h_{t-1}^{(1)}; \Theta^{(1)}) \in \mathbb{R}^{d_h}$$

This produces a sequence of short-term representations $H^{(1)} = \{h_1^{(1)}, \dots, h_L^{(1)}\}$, capturing fine-grained transactional dynamics within each window.

To model longer-term dependencies, a second recurrent module (Enc2) operates over sub-sequence summaries. Specifically, for each sub-sequence $k$, the final hidden state $h_{L_k}^{(1)}$ is used as input to the long-term encoder:

$$h_k^{(2)} = \mathrm{LSTM}(h_{L_k}^{(1)}, h_{k-1}^{(2)}; \Theta^{(2)}) \in \mathbb{R}^{d_h}, k = 1, \dots, K$$

This hierarchical structure allows the model to capture both short-term behavioral patterns (e.g., burst activity, rapid transfers) and longer-term trends (e.g., gradual accumulation, periodic behavior) that span multiple sub-sequences.

### *3.3 Temporal Fusion and Static-Dynamic Context Integration*

In the current implementation, static features are not directly integrated into the temporal encoder. Preliminary experiments indicated that direct injection of static context into the encoder introduced shortcut signals that allowed the contrastive objective to be optimized without relying on temporal dynamics. As a result, the model could distinguish positive and negative pairs based on static attributes rather than sequential behavior.

To mitigate this effect, the present formulation focuses exclusively on dynamic transactional features, ensuring that the learned representations reflect temporal structure rather than entity identity. This design choice enforces that the encoder captures behavioral evolution across sequences rather than static characteristics of entities.

The integration of static features in a controlled manner remains an area for future investigation. Potential directions include conditioning attention mechanisms on static context or using static features to modulate feature selection without directly influencing the representation space used in the contrastive objective.

### *3.4 Contrastive Predictive Coding Objective*

Training follows the Contrastive Predictive Coding (CPC) framework [4]. Let $c_t = \tilde{c}_t$ denote the context vector at time $t$. For each prediction horizon $k$, a linear projection produces:

$$\hat{z}_{t+k} = W_k c_t \in \mathbb{R}^{d_h}$$

Positive pairs are defined as $(\hat{z}_{t+k}, z_{t+k})$, where $z_{t+k}$ is the representation of the true future sub-sequence obtained by encoding the corresponding local window using the short-term encoder (Enc1). In contrast to $\hat{z}_{t+k}$, which is predicted from the global context representation, $z_{t+k}$ is computed directly from observed data and reflects local temporal structure within the future window. This asymmetry ensures that the contrastive objective aligns global context representations with locally encoded future behavior, rather than comparing two representations produced by identical processing paths. The InfoNCE loss is:

$$\mathcal{L}_k = -\mathbb{E}\log\left(\frac{\exp(\hat{z}_{t+k}^{\top} z_{t+k}/\tau)}{\exp(\hat{z}_{t+k}^{\top} z_{t+k}/\tau) + \sum_j \exp(\hat{z}_{t+k}^{\top} z_j^{-}/\tau)}\right)$$

where $\tau > 0$ is a temperature parameter controlling the sharpness of the distribution.

The total loss is:

$$\mathcal{L}_{\mathrm{TCT}} = \frac{1}{K}\sum_{k=1}^{K} \mathcal{L}_k$$

Minimizing this objective maximizes a lower bound on mutual information between the context $c_t$ and future latent states [4,7], encouraging the encoder to retain predictive temporal structure while suppressing noise.

### *3.5 Downstream Classifier*

After self-supervised pre-training, the encoder parameters are frozen. For each party $p$, an embedding is extracted as:

$$e_p = f_\theta(S_p)$$

A supervised gradient-boosted classifier $g_\phi: \mathbb{R}^d \to [0,1]$[16] is trained on labeled data to produce a risk score:

$$r_p = g_\phi(e_p)$$

A transaction is flagged when $r_p > \alpha$, where $\alpha$ is a predefined threshold.

The learned embeddings can also be used in unsupervised anomaly detection methods (e.g., LOF, ECOD [17,18]) to support exploratory analysis of behavioral patterns without requiring labeled data.

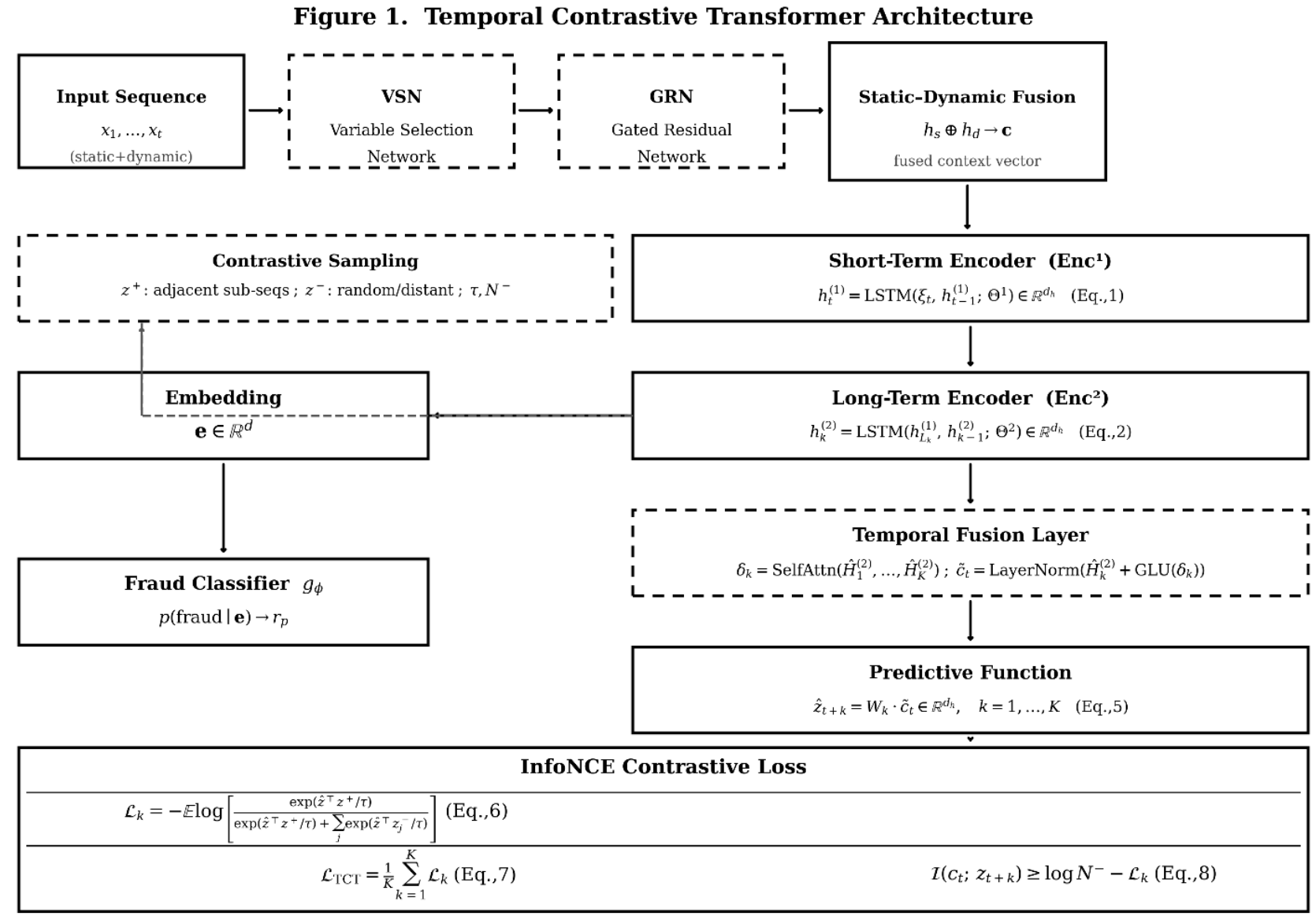

*Figure 1. Temporal Contrastive Transformer architecture. Solid boxes denote primary processing modules; dashed boxes denote auxiliary modules. Arrows indicate data flow during both pre-training and inference. The contrastive objective is active only during pre-training.*

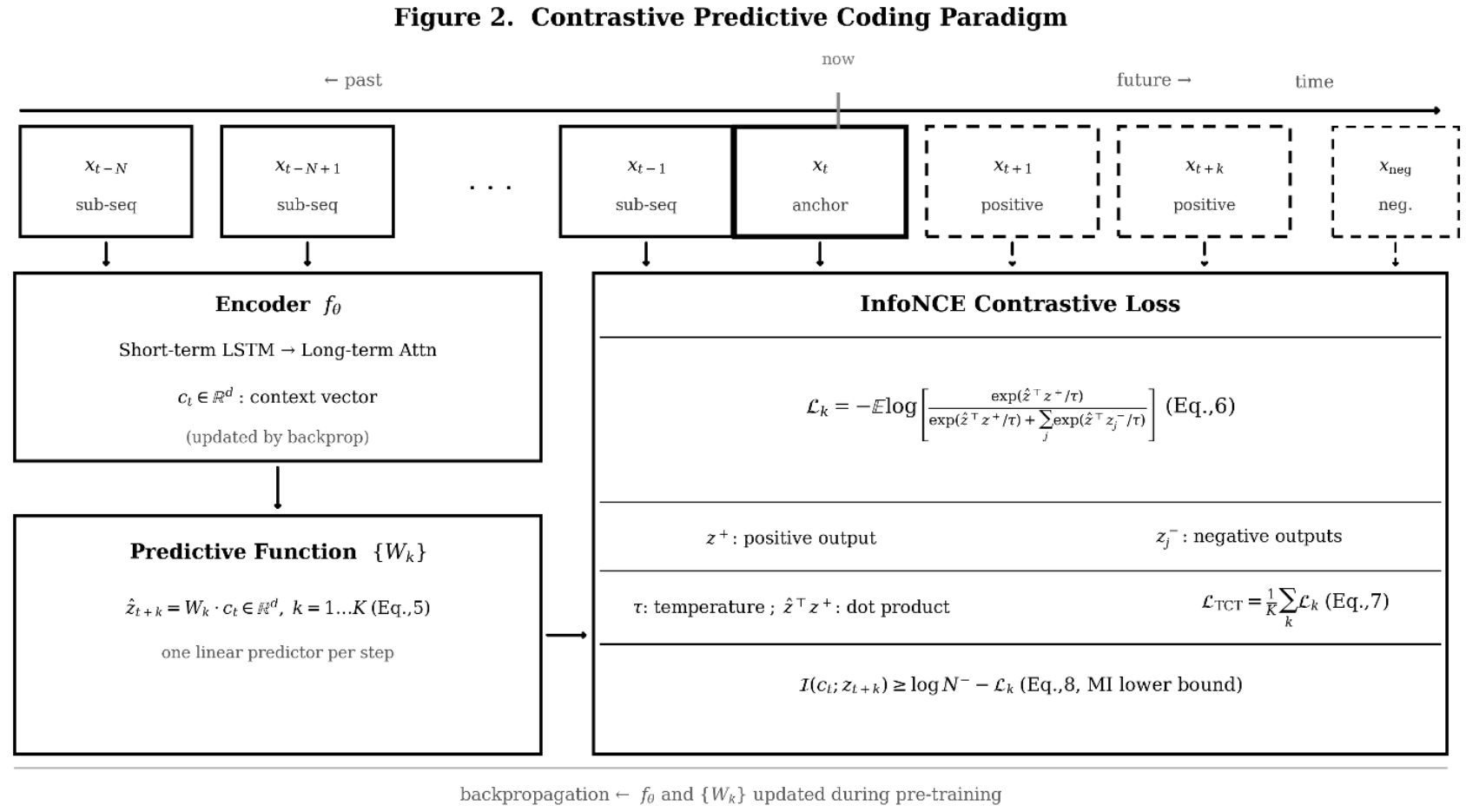

*Figure 2. Contrastive Predictive Coding paradigm within TCT. Past sub-sequences are encoded into a context vector $\tilde{c}_t$. A predictive function $W_k$ projects $\tilde{c}_t$ to a predicted future representation $\hat{z}_{t+k}$. The InfoNCE loss contrasts the prediction against the true positive $z^+$ and a set of negatives $\{z_i^-\}$ sampled from the mini-batch.*

## 4. EXPERIMENTAL SETUP

### *4.1 Dataset and Pre-processing*

Experiments are conducted on a real-world retail payments dataset comprising wire-transfer and electronic transactions associated with business entities. The data are partitioned chronologically into training, validation, and held-out test sets to prevent temporal leakage. The dataset spans multiple years and comprises millions of transactions across a large population of entities, providing sufficient scale for self-supervised pre-training while avoiding disclosure of sensitive operational details.

Feature engineering is intentionally limited. Raw transaction attributes are retained together with a small set of behavioral statistics that can be computed in real time, reflecting operational deployment constraints. These include simple aggregations such as transaction counts, recency measures, and channel-based activity indicators.

Event sequences are constructed at the entity level by ordering all transactions for a given party by timestamp. The sequences are segmented into sub-sequences used as input to the TCT encoder. For self-supervised pre-training, sequences were required to contain at least 15 historical events together with a future window of at least 3 events, corresponding to the minimum local prediction window. For the supervised downstream task, a fixed window length of 3 events was used.

The labeled data exhibit strong class imbalance, with fraudulent entities representing a small minority. Evaluation therefore relies on AUC-ROC, which summarizes performance across all classification thresholds and is insensitive to class prevalence.

### *4.2 Model Configuration*

The TCT encoder is configured with embedding dimension $d = 32$ and LSTM hidden state dimension $d_h = 32$. Sequence length varies between 15 and 60 events. Each sequence is decomposed into 5–6 global sub-sequences, with adaptive local window sizes reflecting transaction density.

Training uses the AdamW optimizer [19] with weight decay $10^{-2}$, a one-cycle learning rate schedule with maximum learning rate $10^{-3}$, and batch size 256, which also determines the number of negative samples in the contrastive objective. The InfoNCE temperature is set to $\tau = 0.1$.

The downstream classifier is an XGBoost model [16] with $n_estimators = 100$, max_depth $= 5$, and learning rate $0.3$, with L1 regularization $\alpha = 0$ and L2 regularization $\lambda = 1$.

### *4.3 Baselines*

We evaluate TCT against two reference configurations. The first is a raw-features baseline, in which XGBoost is trained on 30 expert-engineered behavioral features capturing transaction frequency, recency, novelty of counterparties and devices, and channel-specific activity patterns. This baseline reflects a strong, domain-informed representation commonly used in production systems.

The second configuration augments these features with the TCT embedding vector by concatenation. This setting tests whether the learned representations are hypothesized to provide incremental value beyond the engineered features.

### *4.4 Evaluation*

All models are evaluated on the held-out test set. The primary metric is AUC-ROC. Train and validation AUC values are also reported to assess generalization behavior.
Receiver operating characteristic curves are plotted for all configurations to illustrate performance across the full range of decision thresholds. This is particularly relevant in operational settings, where acceptable false-positive rates vary depending on investigation capacity.

## 5. RESULTS

### *5.1 Quantitative Performance*

Table 1 summarizes AUC-ROC scores across training, validation, and test partitions for all configurations. On the held-out test set, the raw-feature baseline achieves the highest AUC-ROC of 0.9245, followed by the combined raw-features-plus-embeddings configuration at 0.9205, and the embeddings-only configuration at 0.8644.

*Table 1. AUC-ROC summary across experimental conditions.*

| Configuration | AUC-ROC (Train) | AUC-ROC (Val.) | AUC-ROC (Test) |
|---|---|---|---|
| **XGB with raw + embeddings** | 0.992 | 0.951 | **0.9205** |
| XGB without embeddings | 0.977 | 0.937 | 0.9245 |
| XGB with embeddings only | 0.984 | 0.916 | 0.8644 |
| Random (chance) | 0.500 | 0.500 | 0.5000 |

The difference between the raw-feature baseline and the combined configuration is small ($\Delta = -0.004$ AUC), indicating that the addition of TCT embeddings does not yield measurable improvement in this setup. Given the single train/test split, this difference should be interpreted cautiously and does not provide evidence of a systematic degradation in performance. Notably, the raw-feature baseline represents a strong, domain-informed reference point; achieving comparable performance at this stage suggests that the learned representations approximate existing feature abstractions.

The embeddings-only configuration achieves AUC-ROC of 0.8644, demonstrating that the learned representations are trained to capture meaningful temporal signal. However, the gap relative to the raw-feature baseline indicates that these representations are not yet competitive with domain-engineered features when used in isolation.

All models exhibit a noticeable generalization gap between training and test performance (train AUC in the range 0.977–0.992 versus test AUC 0.8644–0.9245), suggesting overfitting. This behavior is more pronounced in the embeddings-based configurations and indicates that further regularization, larger pre-training corpora, or improved training objectives may be required to stabilize representation quality.

### 5.2 ROC Curve Analysis

Figure 3 presents the ROC curves on the held-out test set, with the false-positive-rate (FPR) axis truncated at 0.30 to reflect the operationally relevant range. The curves corresponding to the raw-feature baseline and the combined raw-features-plus-embeddings configuration exhibit near-identical global behavior, consistent with the small difference in AUC (ΔAUC = 0.004).

A more detailed inspection reveals that the curves intersect, indicating that each model is locally superior in different operating regions. In the low-to-mid FPR range (approximately 0.03–0.06), the combined configuration achieves slightly higher true positive rates, while in higher FPR regions the raw-feature baseline marginally dominates. This crossing behavior explains the negligible difference in AUC and indicates that the addition of embeddings modifies the decision boundary without yielding consistent improvement across thresholds.

The embeddings-only configuration exhibits consistently lower performance (AUC 0.8644), confirming that the learned sequential representations capture meaningful signal but do not match the discriminative power of domain-engineered features when used in isolation. At FPR = 0.10, the raw-feature and combined configurations achieve TPR of approximately 0.80, whereas the embeddings-only configuration achieves TPR of approximately 0.69.

Overall, the results indicate that TCT embeddings encode behavioral information that substantially overlaps with engineered features, while introducing localized differences in ranking that are not sufficient to improve aggregate performance under the current configuration. The absence of consistent improvement across thresholds should therefore be interpreted in the context of early-stage representation learning, where parity with engineered features is already a meaningful signal of progress.

### 5.3 Embedding Space Analysis

Cosine similarity analysis of embeddings across local transaction windows reveals a consistent difference in distribution between fraudulent and legitimate activity. Specifically, embeddings associated with fraudulent transactions exhibit higher variability and lower similarity to surrounding context compared to legitimate activity, indicating more abrupt and irregular behavioral patterns. This observation is consistent with the characterization of fraud as anomalous relative to typical transactional behavior, rather than as a stable sequential pattern.

While these results indicate that the embeddings capture meaningful deviations from local behavioral context, the current evaluation shows that this signal does not yet translate into additive predictive value beyond domain-engineered features, suggesting the need for extended pre-training or architectural refinement. This structure is further illustrated in Figure 4. Age is not included in the training objective, indicating that this structure emerges implicitly from temporal behavioral patterns rather than explicit supervision.

**Figure 3. ROC Curve — Test Data**

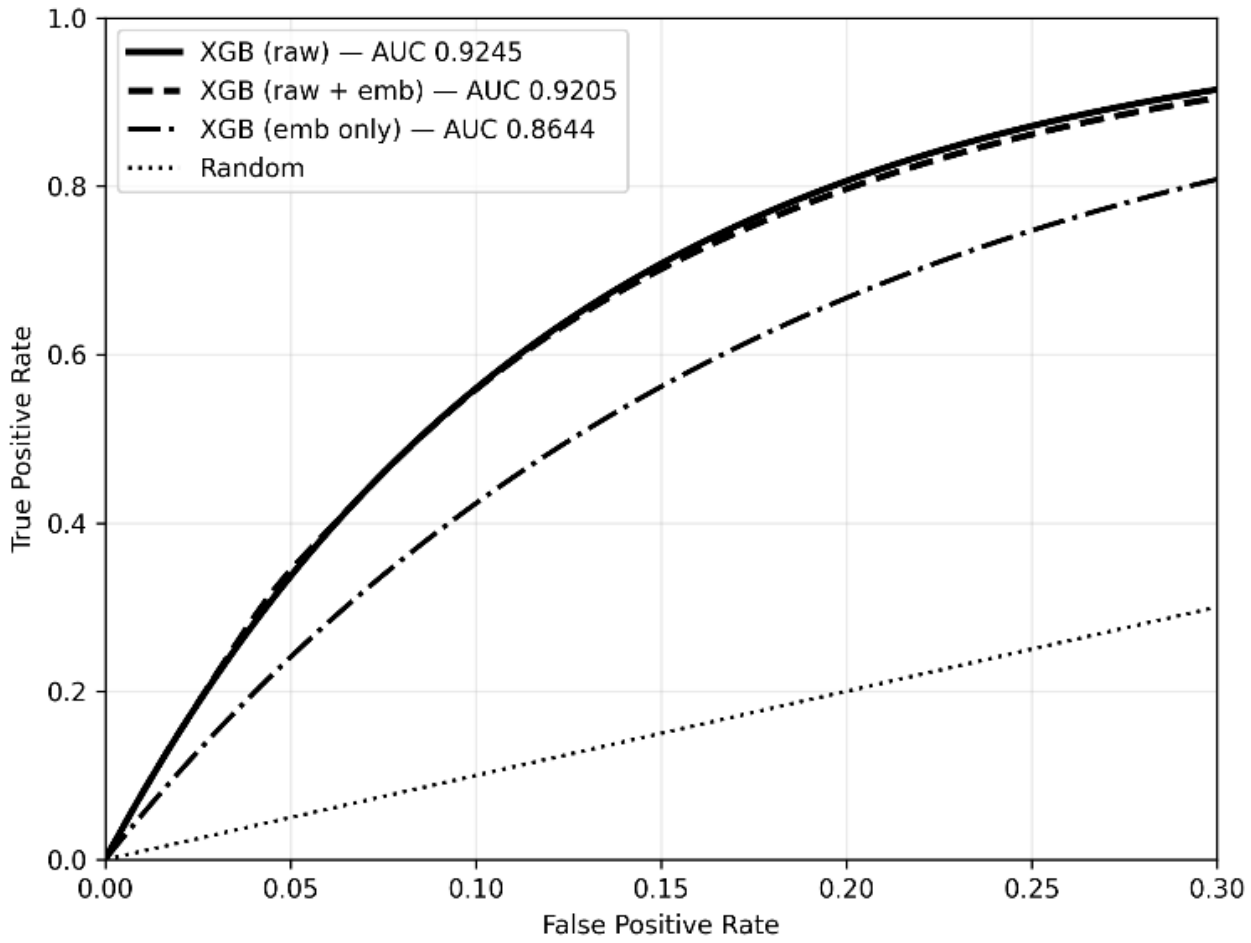

***Figure 3.*** *ROC curves on the held-out test partition (FPR axis truncated at 0.30). The solid curve XGB with raw features, (AUC = 0.9245) and the dashed curve XGB with raw + embeddings, (AUC = 0.9205) exhibit near-identical global behavior ΔAUC = 0.004), with minor local deviations across operating regions. The dash-dotted curve XGB with embeddings only, (AUC = 0.8644) lies substantially below both, indicating lower discriminative performance. The dotted diagonal represents random performance.*

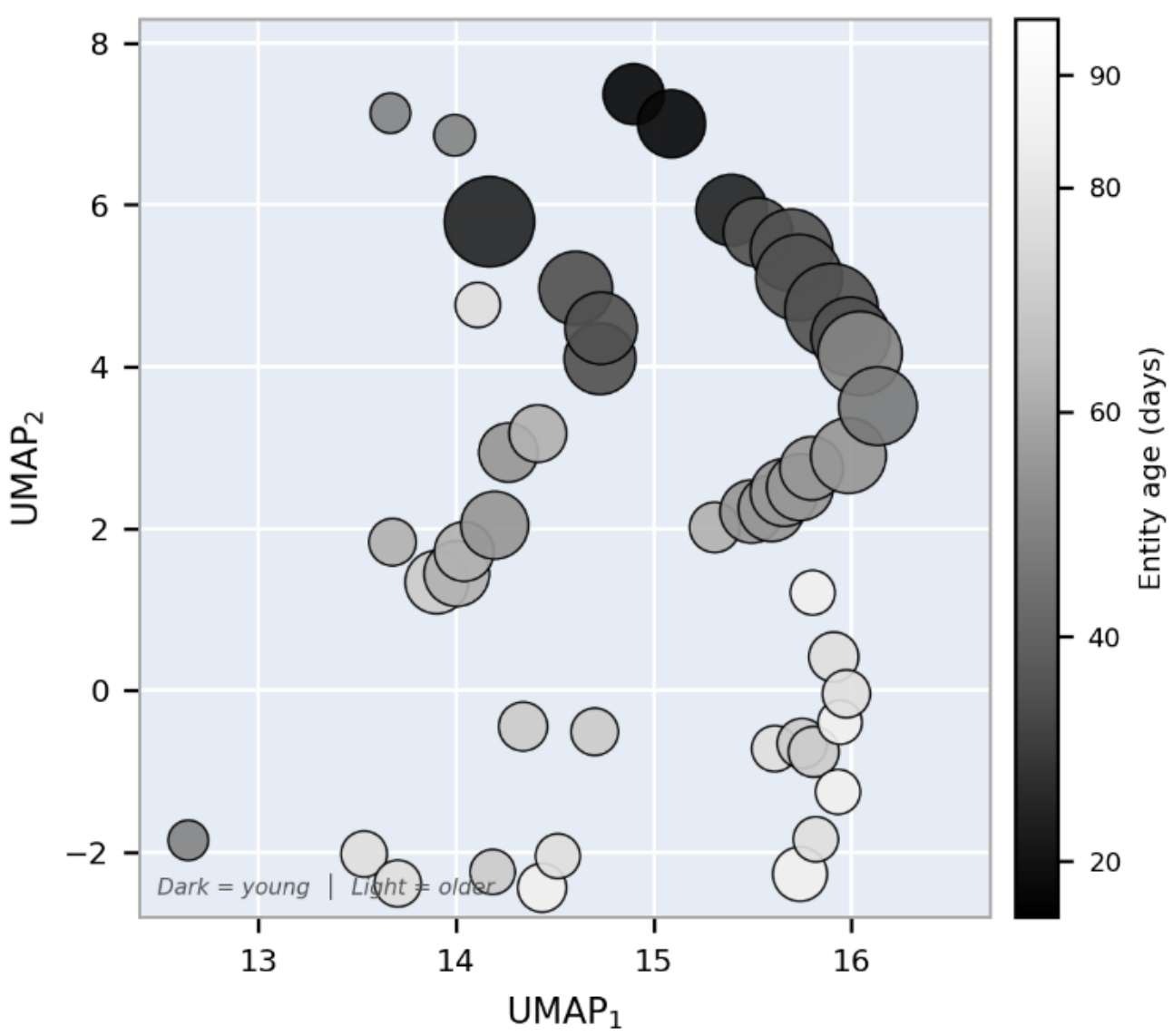

***Figure 4****. UMAP projection of learned embeddings on the validation set. Marker size reflects customer age. Although age is not used as a training signal, the embedding space exhibits smooth age-related structure, with similar age groups forming coherent regions and gradual transitions across space. This structure emerges from temporal behavioral dynamics captured by the model rather than from explicit supervision.*

## 6. DISCUSSION

### *6.1 Generality: Fraud and Anti-Money Laundering*

The TCT framework is conceptually applicable across financial sequence analysis tasks in which anomalous behavior manifests over time. Both fraud detection and AML share a common structural property: suspicious activity is not defined by individual transactions, but by temporal patterns-sequences of events that deviate from typical behavioral trajectories. TCT addresses this by learning a representation in $\mathbb{R}^d$ that is trained to capture temporal dynamics of transaction behavior, enabling downstream models to detect deviations from these patterns.

In AML settings, the availability of labeled data is particularly limited. Suspicious activity reports (SARs) are sparse, subject to reporting bias, and often incomplete with respect to the full transactional context. In this setting, self-supervised pre-training provides a practical advantage: the encoder can be trained on the full unlabeled transaction corpus and subsequently applied in either supervised or unsupervised modes. In the supervised case, embeddings can complement domain features; in the unsupervised case, anomalies in embedding space may serve as signals for investigation.

However, the empirical results in this study indicate that, under the current configuration, embeddings do not provide measurable additive value over strong domain-engineered features. Rather than indicating failure, this outcome reflects that the learned representations are trained to capture behavioral signal already encoded by engineered features. At this stage, achieving comparable performance to a strong baseline is itself a meaningful result, suggesting that representation learning can approximate domain-specific feature abstractions. This points to a promising direction toward reducing reliance on manual feature engineering, although the approach is not yet ready for production deployment.

### *6.2 Why Self-Supervised Pre-Training Helps*

The InfoNCE objective [4,7] introduces a predictive inductive bias by training the encoder to infer future latent representations from past context, encouraging the model to capture temporal structure in transaction sequences. This enables representation learning to be decoupled from supervised classification, allowing the encoder to leverage large volumes of unlabeled data and learn general behavioral patterns prior to task-specific adaptation. Empirically, the embeddings achieve meaningful predictive performance in isolation, indicating that they encode non-trivial temporal signal. However, the absence of improvement when combined with domain-engineered features suggests that the current formulation primarily recovers information already captured by existing behavioral aggregates. The key challenge, therefore, is not the ability to learn representations, but to learn representations that capture information not already expressed through engineered features.

### *6.3 Limitations and Future Directions*

Several limitations warrant discussion. First, the current formulation models each party independently as a temporal sequence and does not explicitly incorporate relational structure across entities, such as shared devices, counterparties, or multi-hop transaction paths. We observed that direct integration of static features into the encoder can introduce shortcut signals in the contrastive objective, allowing the model to distinguish samples without relying on temporal dynamics. This highlights the need for controlled integration mechanisms when combining static and sequential information. Graph-based approaches have demonstrated the importance of such structure in AML contexts [9,13], and integrating relational representations with TCT embeddings is a natural extension.

Second, the effectiveness of contrastive learning depends on the quality of negative samples. The current implementation uses uniformly sampled in-batch negatives ($N^- = 256$), which may not provide sufficiently challenging contrasts for subtle behavioral distinctions. Incorporating hard negative mining strategies, as explored in MoCo [5] and SimCLR [6], may improve the discriminative power of the learned representations.

Third, the downstream classifier may limit the utilization of embedding information. The current XGBoost configuration is intentionally shallow and optimized for efficiency; however, more expressive models or end-to-end fine-tuning may better exploit the structure encoded in the embeddings.

Finally, the computational complexity of the architecture is influenced by the self-attention component, which scales with the number of sub-sequences. Approximations such as sparse or linear attention may be required to scale the model to very long transaction histories.

## 7. CONCLUSION

We introduced the Temporal Contrastive Transformer (TCT), a self-supervised framework for learning representations of financial transaction sequences. By combining a Temporal Fusion Transformer backbone [3] with a Contrastive Predictive Coding objective [4], TCT learns embeddings that capture temporal behavioral structure without relying on labeled data.

Empirical evaluation on a commercial payment fraud dataset shows that the learned embeddings encode meaningful signal (AUC 0.8644 in isolation). However, when combined with domain-engineered features, they do not yield measurable improvement over a strong baseline (AUC 0.9205 vs. 0.9245), indicating that the current representations largely overlap with existing feature abstractions.

These findings highlight both the promise and current limitations of contrastive sequential pre-training for financial crime detection. The learned representations capture meaningful signal and achieve performance comparable to strong domain-engineered features, indicating a viable direction for reducing reliance on manual feature engineering. However, the absence of measurable performance gains at this stage suggests that further development is required before production deployment. Future work will focus on scaling pre-training data, extending temporal context, and refining training objectives to unlock additive value beyond engineered features.